\documentclass{article}

\usepackage[table,dvipsnames]{xcolor}

\usepackage{multirow}

\usepackage{amsmath}
\usepackage[english]{babel}

\usepackage{listings}

\usepackage[preprint]{neurips_2024}

\usepackage[T1]{fontenc}    
\usepackage[utf8]{inputenc} 
\usepackage{textcomp} 
\usepackage{hyperref}       
\usepackage{url}            
\usepackage{booktabs}       
\usepackage{amsfonts}       
\usepackage{nicefrac}       
\usepackage{microtype}      
\usepackage{xcolor}         
\usepackage{graphicx}
\usepackage{subcaption}

\title{Teaching LLMs to Refine with Tools}

\definecolor{TableShade}{gray}{0.96}
\definecolor{TableVividGreen}{RGB}{0, 196, 0 }
\definecolor{TableGreen}{RGB}{243,247,240}
\definecolor{TableRed}{RGB}{220, 20, 60}
\newcommand{\tc}[1]{\textcolor{TableVividGreen}{#1}}
\newcommand{\tr}[1]{\textcolor{TableRed}{#1}}
\usepackage{amssymb}
\usepackage{pifont}
\newcommand{\cmark}{\ding{51}}%
\newcommand{\xmark}{\ding{55}}%
\author{Dian Yu, Yuheng Zhang, Jiahao Xu, Tian Liang, Linfeng Song, \\ \bf Zhaopeng Tu, Haitao Mi, and Dong Yu \\
Tencent AI Lab\\
\texttt{\{yudian,yuhenyzhang,haitaomi,dyu\}@global.tencent.com} \\
}

\begin{document}

\maketitle

\begin{abstract}

Large language models (LLMs) can refine their responses based on feedback, enabling self-improvement through iterative training or test-time refinement. However, existing methods predominantly focus on refinement within the same reasoning format, which may lead to non-correcting behaviors. We propose CaP, a novel approach that uses external tools to refine chain-of-thought (CoT) responses generated by the same or other LLMs. CaP employs a two-stage training process: supervised fine-tuning followed by preference optimization with DPO variants. Our observations highlight the critical role of preference optimization in enabling effective refinement. Additionally, we compare several sampling strategies to leverage CoT and tools at inference time. Experimental results demonstrate CaP's potential for effective cross-reasoning refinement and efficient inference.

\end{abstract}

\section{Introduction}

\begin{figure*}[h!]
   \begin{center}
   \includegraphics[width=0.88\textwidth]{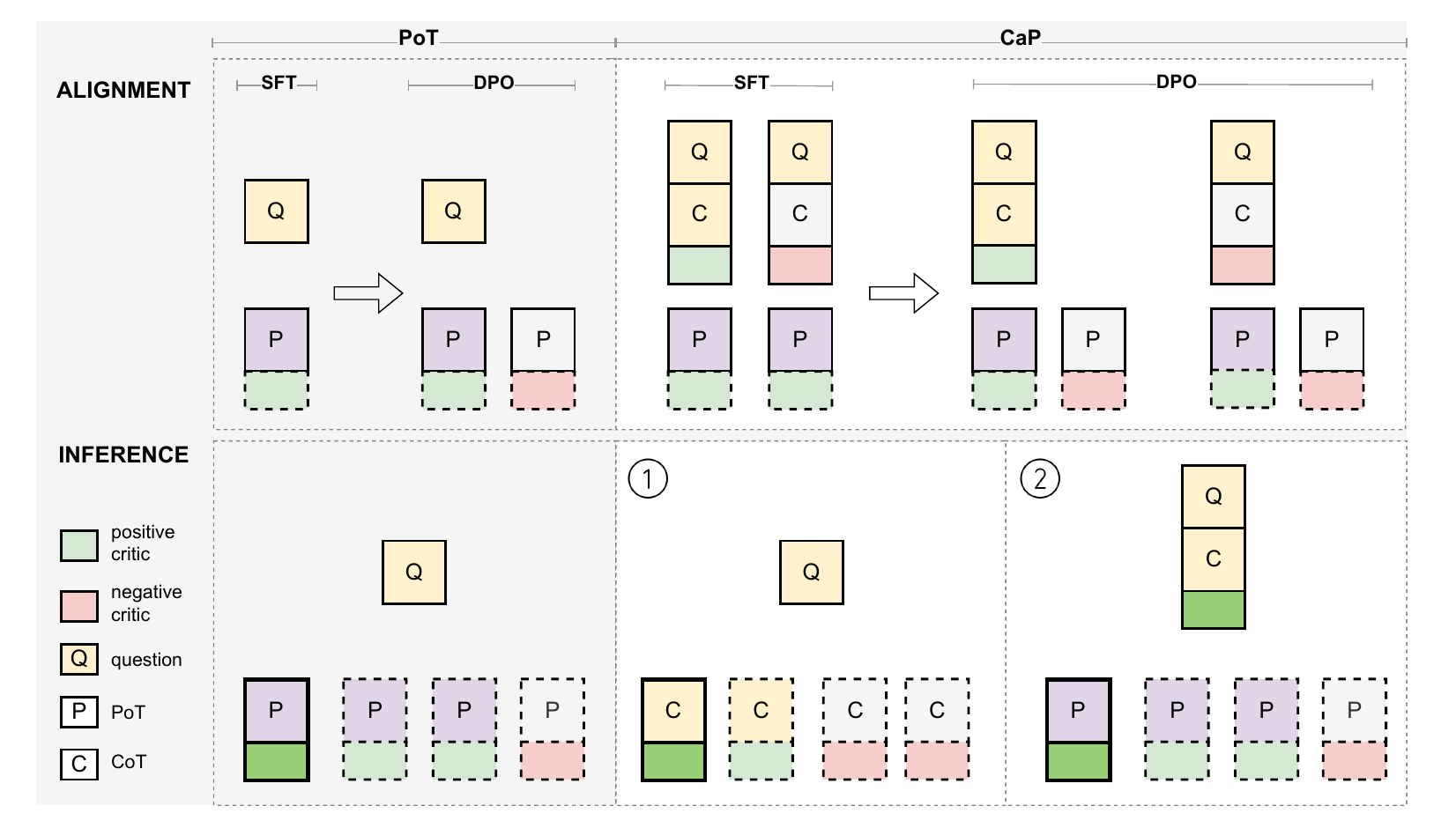}
   \end{center}
 \caption{Overview of refining CoT solutions with PoT solutions during alignment and inference.  }
 \label{fig:overview}
\end{figure*}

It has been shown that large language models (LLMs) can exhibit certain capabilities of refining their responses based on feedback~\citep{saunders2022self,madaan2024self,qu2024recursive,kumar2024training}, thereby enabling self-improvement of LLMs via iterative training or at test-time. Recent studies show that simply prompting LLMs can hardly achieve effective refinement even when the reference answer is provided along with the previous attempt, and thus further alignment such as supervised fine-tuning and preference optimization is still essential~\citep{qu2024recursive,kumar2024training}. Most studies focus on response-level refinement, likely because obtaining large-scale, high-quality process-level supervision at a low annotation or computation cost~\citep{lightman2023let,wang2023shepherd} remains an open research question.

A key step is constructing response pairs, consisting of a previous attempt and its refined version. To scale this process without incurring high annotation costs, previous studies often sample responses from LLMs in a multi-turn fashion, followed by answer verification and filtering: reference answers and a prior failed attempt are provided to guide the generation of a successful attempt~\citep{zelikman2022star}. Alternatively, correct and incorrect responses can be paired to simulate the multi-turn process~\citep{welleck2022generating,snell2024scaling}. However, these studies focus on scenarios where responses follow the \textbf{same} reasoning format, such as step-by-step reasoning in natural language (CoT~\citep{wei2022chain}) or solving problems through executable code (PoT~\citep{chen2022program}).

Leveraging different types of reasoning~\citep{gou2024tora,yue2024dots} has proven effective, particularly for improving LLMs' performance in mathematical reasoning. Responses of different reasoning types for the same question naturally serve as alternative solutions, given the inherent differences between programming language and natural language. However, the problem of refining responses across distinct types of reasoning remains underexplored. As a preliminary investigation, we focus on  the question: \textbf{Can we teach LLMs to refine CoT responses using tools?}

\begin{figure}[ht]
    \centering
    \begin{subfigure}{0.49\textwidth}
        \centering
        \includegraphics[width=\textwidth]{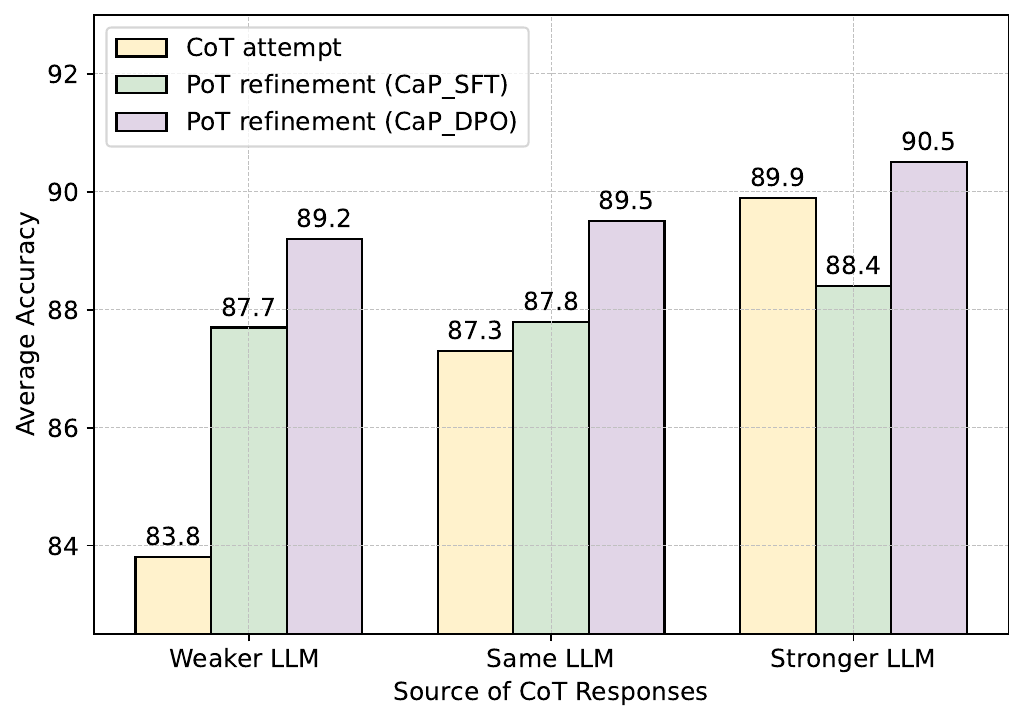}
        \label{fig:left}
    \end{subfigure}
    \hspace{-0.1cm}
    \begin{subfigure}{0.49\textwidth}
        \centering
        \includegraphics[width=\textwidth]{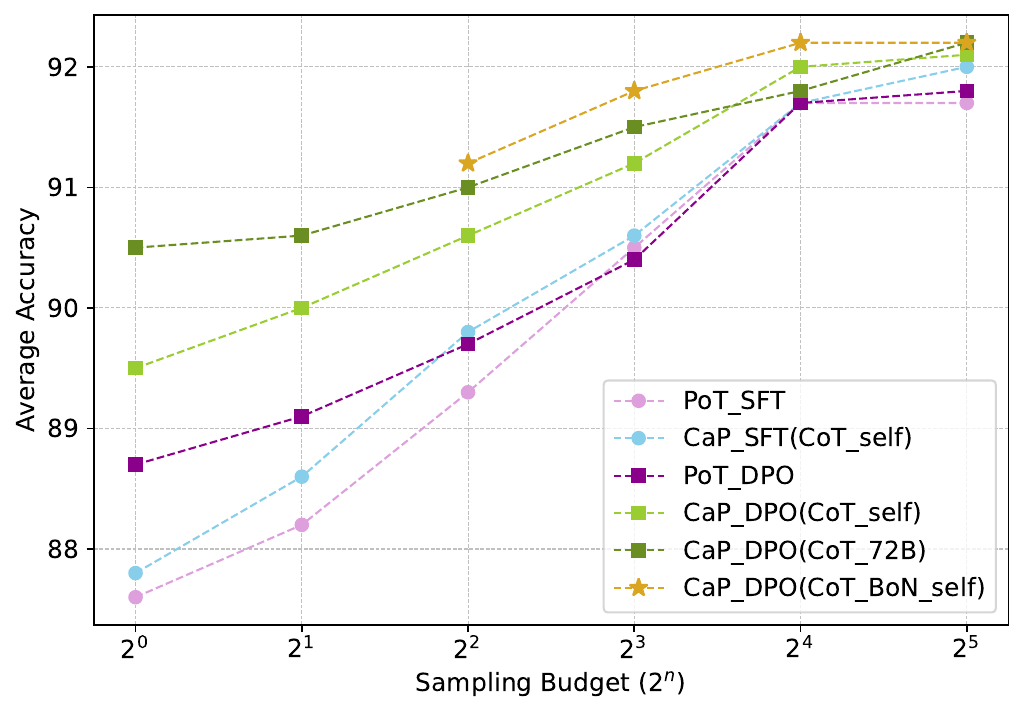}
        \label{fig:right}
    \end{subfigure}
    \caption{Left: CaP performance using greedy decoding based on different sources of CoT responses. Right: Average accuracy of BoN and BoNBoN sampling strategies on out-of-distribution math tasks.}
    \label{fig:combined}
\end{figure}

We propose an approach called \textbf{CaP}, which leverages external tools to effectively refine CoT responses, regardless of whether CoT responses are generated by the same LLM, a stronger one, or a weaker one. In turn, CoT responses can also serve as context or draft information to facilitate the generation of PoT solutions. 
To develop CaP, we train LLMs through a two-stage process: supervised fine-tuning (SFT) followed by preference optimization using DPO~\citep{rafailov2024direct} as shown in Figure~\ref{fig:overview}. Our observations indicate that SFT alone is insufficient to instill effective refinement behavior; the subsequent preference optimization stage is essential for enabling this capability (left subfigure in Figure~\ref{fig:combined}). When applying the same paradigm as CaP but using CoT for refinement, we observe negative results, aligning with prior studies~\citep{kumar2024training,ye2024physics}, which reveals that LLMs trained to refine CoT with CoT are likely to exhibit non-correcting behaviors.

 Finally, we explore the benefits of CaP during inference. Compared to the PoT/CoT-only variants, our CaP model consistently achieves superior Best-of-N (BoN) performance within the same sample budget, and it further narrows the gap between one-sample and BoN performance (Figure~\ref{fig:combined}), an essential step toward reducing the computational overhead during inference~\citep{sessa2024bond}. Recognizing the importance of CoT response correctness for refinement performance, we compare several sampling strategies and introduce a new sampling strategy called \textbf{BoNBoN}, which allocates half of the budget to CoT sampling and then uses the resulting BoN CoT sample as the initial attempt to generate N refined samples in PoT for BoN selection. The positive results suggest a promising direction for studying adaptive allocation of inference-time computation by leveraging diverse reasoning steps or solutions for efficient inference.

\section{Method}

\subsection{Best-of-N Sampling from Teacher Models}
\label{sec:method:bon}

For complex tasks such as mathematical reasoning, evaluating the correctness of a response based solely on the generated rationale remains a significant challenge~\citep{daheim2024stepwise}. Therefore, to construct high-quality training data, we focus on a setting where a reference answer -- usually concise and lacking any form of rationale -- is provided to assist the evaluation. We apply a reference-based critic model, trained with next-token prediction to evaluate and score each CoT response \( y_{\text{cot}} \) or each PoT response \( y_{\text{pot}} \) and its corresponding execution result \( y_{\text{exec}} \), given a question \( q \) and its ground truth reference answer \( y_{\text{ref}} \). We use the aforementioned critic model to generate \textsc{Yes} or \textsc{No} critics and regard the likelihood that a solution is positive or negative as the confidence score for a certain critic (e.g., $p_{\theta_{\text{ref}}}(\text{Yes} \mid q, y_{\text{pot}}, y_{\text{exec}}, y_{\text{ref}})$ and $p_{\theta_{\text{ref}}}(\text{No} \mid q, y_{\text{cot}}, y_{\text{ref}})$), inspired by previous studies on reference-free/reference-based outcome/process reward modeling~\citep{lightman2023let,zhang2024generative, tian2024toward}. While more detailed critics that pointing out errors can undoubtedly enhance both evaluation and refinement, they heavily depend on human-annotated data~\citep{mcaleese2024llm}, which we leave for future work.

Supposing that we have two teacher policy models $\pi_{\text{cot}}$ and $\pi_{\text{pot}}$, we use Best-of-N sampling strategy to build SFT and preference optimization training data. We draw $N_1$ CoT samples from $\pi_{\text{cot}}$ and $N_2$ PoT samples from $\pi_{\text{pot}}$. We use the highest-scoring responses as the winning and losing responses ($y_\text{cot}^{+}$, $y_\text{cot}^{-}$, $y_\text{pot}^{+}$, $y_\text{pot}^{-}$) in later stages, to reduce noise that may be introduced by an imperfect critic model. Note that a question may have only positive (+) or negative (-) responses, depending on the evaluations of the critic model.

Instead of prompting LLMs in a multi-turn manner, we explore pairing CoT responses with positive PoT responses to simulate a multi-turn process. This is inspired by previous studies on CoT-CoT refinement~\citep{welleck2022generating,snell2024scaling}. Refining a CoT response using another CoT response, even when reference answers are available, remains an open challenge for LLMs~\citep{zelikman2022star} and leveraging external tools for refinement introduces additional complexities.

\subsection{Training a Reference Policy Model}
\label{sec:method:sft}

Let $\pi(x, \cdot)$ be a policy that generates sequences $y$ given prompt $x$. We define  $c^{+} =$ ``\textsc{The problem-solving process might be correct.}'', and $c^{-}=$ ``\textsc{The problem-solving process might be wrong.}'', corresponding to the critics \textsc{Yes} and \textsc{No}, respectively. 

One natural choice is to perform SFT on the Best-of-N PoT and CoT responses in the instruction tuning data. We use this unified training objective to enhance the ability of LLMs to generate accurate single-turn CoT responses and enabling the refinement of CoT responses through PoT reasoning.

\begin{equation}
\footnotesize
D_{\text{CaP}} = \left\{ \left( q, y_{\text{cot}}^{+}, c^{+}, y_{\text{pot}}^{+} \right) \right\} 
\cup 
\left\{ \left( q, y_{\text{cot}}^{-}, c^{-}, y_{\text{pot}}^{+} \right) \right\}
\end{equation}

\begin{equation}
\small
D_{\text{CoT}} = \{ (q, y_{\text{cot}}^{+}) \} 
\end{equation}

\begin{equation}
\small
L_{\text{SFT}}(\pi_{\theta}) = - \sum_{D_{\text{CaP}}} \log \pi_{\theta} \left( y_{\text{pot}}^{+} \mid q, y_{\text{cot}}, c \right) - \sum_{D_{\text{CoT}}} \log \pi_{\theta} \left( y_{\text{cot}}^{+} \mid q \right)
\end{equation}

\subsection{Preference Optimization}
\label{sec:method:dpo}

Initially, we explore preference optimization to utilize the negative PoT responses ($y_\text{pot}^{-}$). We later observe that this stage plays a critical role in encouraging refinement behavior (Section~\ref{sec:experiment}). Additionally, we apply a variant of DPO~\citep{rafailov2024direct} that incorporates an SFT loss on the Best-of-N postive PoT samples for stabilizing preference learning ~\citep{zheng2024llamafactory,xu2024chatglm,pang2024iterative,liu2024provably} and controlling response length. Notably, we focus solely on the multi-turn setting at this stage to bias the implicit reward towards refinement behavior, as the single-turn problem-solving reasoning already been enhanced during the SFT stage.

\begin{equation}
\footnotesize
D_{{\text{CaP}}_{\text{pref}}} = \left\{ \left( q, y_{\text{cot}}^{+}, c^{+}, y_{\text{pot}}^{+}, y_{\text{pot}}^{-} \right) \right\} 
\cup 
\left\{ \left( q, y_{\text{cot}}^{-}, c^{-}, y_{\text{pot}}^{+}, y_{\text{pot}}^{-} \right) \right\}
\end{equation}

\begin{equation}
\centering
\footnotesize
L_{\text{DPO}}(\pi_{\theta}) = - \sum_{D_{{\text{CaP}}_{\text{pref}}}} \left( \log \sigma \left( \beta \log \frac{\pi_{\theta}(y_{\text{pot}}^{+} \mid z)}{\pi_{\theta_{\text{ref}}}(y_{\text{pot}}^{+} \mid z)} - \beta \log \frac{\pi_{\theta}(y_{\text{pot}}^{-} \mid z)}{\pi_{\theta_{\text{ref}}}(y_{\text{pot}}^{-} \mid z)} \right) + \lambda \cdot \log \pi_{\theta}(y_{\text{pot}}^{+} \mid z) \right)
\end{equation}

\noindent where \( z \) is defined as either \( (q, y_{\text{cot}}^{+}, c^{+}) \) or \( (q, y_{\text{cot}}^{-}, c^{-}) \).

We utilize both correct-to-correct and incorrect-to-correct pairings of CoT/PoT responses to improve the robustness of our trained CaP models in refining CoT responses of varying quality (see examples in Table~\ref{tab:multi_turn_example}). Detailed results in Section~\ref{sec:experiment} demonstrate the benefits of incorporating both positive and negative CoT as context.

\begin{table*}[h!]

\centering

\scriptsize

\begin{tabular}{p{1cm}p{12cm}}
\toprule

\multicolumn{2}{c}{\textbf{Turn I}} \\
\midrule
\textbf{Instruction} & Please solve the following problem step by step.  \\

\textbf{Input}  & \begin{lstlisting}[basicstyle=\ttfamily\tiny, breaklines=true, aboveskip=0pt, belowskip=0pt]
At 9:30 on a clock, the acute angle formed between the hour hand and the minute hand is
( ) A. 75° B. 85° C. 95° D. 105°\end{lstlisting}\\
\midrule
\textbf{Output}  & \begin{lstlisting}[basicstyle=\ttfamily\tiny, breaklines=true, aboveskip=0pt, belowskip=0pt] 
First, we need to understand two geometric terms: supplementary angles and complementary angles.
Two angles are supplementary if their sum equals 180°.
Two angles are complementary if their sum equals 90°.
The problem states that the supplementary angle of a given angle is 115°, which means this angle plus 115° equals 180°. To find the angle, we subtract 115° from 180°:
180° - 115° = 65°
Thus, the angle is 65°.
Next, we determine the complementary angle of this angle. Since an angle and its complementary angle add up to 90°, we subtract 65° from 90° to find its complement:
90° - 65° = 25°
Therefore, the complementary angle is 25°, and option D is correct.\end{lstlisting}\\
\midrule

\multicolumn{2}{c}{\textbf{Turn II}} \\
\midrule
\textbf{Instruction} & Given a problem, please provide Python code to solve it based on the reference thought process and its evaluation (both of which may contain errors). \\
\textbf{Input} & \begin{lstlisting}[basicstyle=\ttfamily\tiny, breaklines=true, aboveskip=0pt, belowskip=0pt] 
### QUESTION
At 9:30 on a clock, the acute angle formed between the hour hand and the minute hand is
( ) A. 75° B. 85° C. 95° D. 105°
### REFERENCE THOUGHT PROCESS
First, we need to understand two geometric terms: supplementary angles and complementary angles.
Two angles are supplementary if their sum equals 180°.
Two angles are complementary if their sum equals 90°.
The problem states that the supplementary angle of a given angle is 115°, which means this angle plus 115° equals 180°. To find the angle, we subtract 115° from 180°:
180° - 115° = 65°
Thus, the angle is 65°.
Next, we determine the complementary angle of this angle. Since an angle and its complementary angle add up to 90°, we subtract 65° from 90° to find its complement:
90° - 65° = 25°
Therefore, the complementary angle is 25°, and option D is correct.
### CRITIC
The problem-solving process might be correct.
\end{lstlisting}\\
\midrule
\textbf{Output} & \begin{lstlisting}[basicstyle=\ttfamily\tiny, breaklines=true, aboveskip=0pt, belowskip=0pt]
from sympy import symbols, Eq, solve

# Define variables
a, b = symbols('a b')

# Angle traveled by the minute hand
# Angle traveled by the minute hand in 30 minutes
b = 30 * 6  

# Angle traveled by the hour hand
# Angle traveled by the hour hand at 9 and a half hours
a = 9 * 30 + 30 * 0.5  

# Calculate the acute angle
# Calculate the angle difference between the hour hand and the minute hand
acute_angle = abs(a - b)  

# Return the result
result = {'a': a, 'b': b, 'acute_angle': acute_angle}
result
\end{lstlisting}\\
\bottomrule
\end{tabular}
\caption{Example instances of the multi-turn CaP data.}
\label{tab:multi_turn_example}
\end{table*}

\section{Experiments}
\label{sec:experiment}

\subsection{Data Construction and Implementation}
We use Qwen2-72B-Instruct~\citep{yang2024qwen2} to generate CoT responses and SIaM8B~\citep{yu2024siam} to generate PoT responses. We use the initial model of SIaM8B based on Llama3-8B-Instruct as it is not trained with any training data of Chinese benchmarks. Models that integrate both PoT and CoT~\citep{gou2024tora, wang2023mathcoder,yang2024qwen2} are not suitable for generating refinement data, as the reasoning subparts in each type are incomplete for solving the problem. 
For each question, we sample $N_1=5$ responses from SIaM8B and $N_2=3$ responses from Qwen2-72B-Instruct.
To ensure diversity and quality of data, we use a dataset comprising one million Chinese question-answer pairs  collected under authorized licenses from educational websites. This dataset spans a broad range of educational levels. However, the answers are typically short phrases, which pose challenges for constructing CoT reasoning. As a result, we rely on Qwen2-72B-Instruct instead. Through sampling, validation, and filtering, we construct 1.5M instruction-tuning instances, comprising 0.8M multi-turn data and 0.7M single-turn CoT data ($D_{\text{CaP}}$ and $D_{\text{CoT}}$ as described in Section~\ref{sec:method:sft}), and 355K preference-pair instances ($D_{{\text{CaP}}_{\text{pref}}}$ detailed in Section~\ref{sec:method:dpo}).

We train two types of critic models-reference-based and reference-free-to generate training data for CaP and to perform Best-of-N selection during inference. As described in Section~\ref{sec:method:bon}, these critic models are trained using next-token prediction loss. To simplify the critic tasks, these models are designed to assess the correctness of a single response, either in CoT or PoT formats, without considering information from previous turns (see examples in Table~\ref{tab:reference_free_critic_cot}, Table~\ref{tab:reference_free_critic_pot} and Table~\ref{tab:reference_based_critic_pot}). For creating reference-based training data, we use GPT-4-0613 to annotate approximately 30K training instances. The trained reference-based model is then used to label CoT/PoT samples. Finally, we  combine the resulting pseudo-labeled data with the GPT-labeled data ($\sim$2M in total) to train a reference-free critic model. Notably, incorporating code execution results as context did not yield significant improvements in the Best-of-N performance when using the reference-free critic model. To improve efficiency, we execute only the selected Best-of-N responses during sampling evaluation. We use Qwen2-7B-Instruct~\citep{yang2024qwen2} as the backbone model for main experiments.

\subsection{Evaluation Datasets}
Since the training data is in Chinese, we evaluate all methods on three Chinese mathematical benchmarks: CM17K~\citep{qin2021neural}, APE~\citep{zhao2020ape210k}, and CMATH~\citep{wei2023cmath}, without utilizing their training data. Our methods, however, are easily adaptable to other languages. Performance is reported in accuracy.

\begin{table*}[h!]
\centering
\footnotesize
\begin{tabular}{llllllllll}
\toprule
model      & history   & tool & \multicolumn{2}{c}{CM17K}  & APE & \multicolumn{2}{c}{CMATH} &  \multicolumn{2}{c}{average} \\
                        & CoT &  & valid & test & valid & valid & test &  \\
\midrule

Qwen2-72B-Instruct & --  & \xmark & 88.7 & 90.9 & 81.9 & 92.8 & 95.0 & 89.9\\
Qwen2-7B-Instruct & --  & \xmark & 83.3  & 82.6 & 76.5 & 86.2 & 90.6 & 83.8\\
SIaM(Llama3-8B-Instruct) & --  & \cmark & 83.3  & 84.7 & 83.2 & 87.2 & 88.5 & 85.4 \\
\midrule

DOTS                 &  -- &  mixed & 86.1 & 85.4 & 82.1 & 90.2 & 90.6 & 86.9 \\
\rowcolor{TableGreen}
SiAM$_\text{SFT}$    &  -- &  \cmark & 87.0 & 86.9 & 83.8 & 89.5 & 91.1 & 87.6 \\
SiAM$_\text{DPO}$    &  -- &  \cmark & 88.3 & 88.0 & 84.3 & 90.7 & 92.4 & 88.7 \\
\rowcolor{TableGreen}

Pair$_\text{SFT}$                & --   & \xmark & 86.5 & 86.6 & 79.4  & 89.2 & 92.3  & 86.8 \\       
\rowcolor{TableGreen}
                    & CoT$_\text{7B}$     & \cmark  & 85.7 & 85.3 & 79.3 & 89.2 & 92.7 & 86.4 \\
\rowcolor{TableGreen}
                   & CoT$_\text{self}$    & \cmark   & 86.1 & 86.2 & 78.9 & 90.2 & 92.8 & 86.8  \\

\rowcolor{TableGreen}
 & CoT$_\text{72B}$    & \cmark  & 85.6 & 88.0 & 79.4 & 89.3 & 93.2 & 87.1\\

Pair$_\text{DPO}$    & --   & \xmark & 86.6 & 86.8 & 79.3  & 89.2  & 92.0   & 86.8  \\      
                  & CoT$_\text{7B}$     & \cmark  & 86.5  & 86.4 & 79.0 & 90.0 & 93.1  & 87.0  \\
                  & CoT$_\text{self}$    & \cmark   & 86.1 & 85.9 & 78.8 & 89.8 & 92.6  & 86.7  \\

                   & CoT$_\text{72B}$  & \cmark  & 85.8 & 87.5 & 79.2  & 89.5 & 92.2 & 86.8\\

\midrule
\rowcolor{TableGreen}
CaP$_\text{SFT}$       & --   & \xmark & 86.6 & 87.5 & 79.4 & 89.7 & 93.2  & 87.3  \\ 
\rowcolor{TableGreen}
                   & CoT$_\text{7B}$     & \cmark  & 87.0 & 87.1 & 83.6 & 89.8 &  91.1 & 87.7  \\
\rowcolor{TableGreen}
                   & CoT$_\text{self}$    & \cmark   & 86.9 & 87.1 & 84.3 & 89.8 & 91.1  & 87.8  \\
\rowcolor{TableGreen}

                     & CoT$_\text{72B}$    & \cmark  & 87.4 & 88.6 & 84.7  & 90.0 & 91.2 & 88.4\\

CaP$_\text{DPO}$   & --   & \xmark & 86.4 & 87.4 & 79.7 & 90.0 & 92.3  & 87.2  \\
                    & CoT$_\text{7B}$     & \cmark  & 87.7 & 88.1 & 85.3 & 90.8 & 93.9 & 89.2 \\
                   & CoT$_\text{self}$    & \cmark   & 88.3 & 89.3 & 85.5 & 90.8 & 93.6  & 89.5  \\
             
                     & CoT$_\text{72B}$    & \cmark  & 88.8 & 90.7 & 86.0 & 92.7 & 94.2 & 90.5 \\

\bottomrule
\end{tabular}
\caption{Accuracy on out-of-distribution Chinese mathematical reasoning benchmarks using greedy decoding. During inference, a positive critic is consistently applied to all CoT attempts. CoT$_\text{7B}$ and CoT$_\text{72B}$ refer to the CoT responses generated by Qwen2-7B-Instruct and Qwen2-72B-Instruct, respectively. CoT$_\text{self}$ denotes the self-generated CoT attempt. }

\label{tab:main_results}
\end{table*}

\subsection{Main Results}

We use greedy decoding for all evaluations, except for scaling test-time experiments. Since we generate only a single CoT attempt without requiring Best-of-N selection, we consistently apply a positive critic to all CoT attempts to provide context for refinement.

We compare CaP with several methods (or their variants): Pair-SFT~\citep{welleck2022generating,kumar2024training}, which uses CoT for refinement; Pair-DPO: following the idea of Pair-SFT, we use the fine-tuned Pair-SFT model as the reference policy for another round of DPO similar to our CaP setting; DOTS~\citep{yue2024dots}, which enables LLMs to dynamically select CoT or PoT reasoning to solve a problem (excluding the analysis layer that may involve question decomposition and rewriting); and SIaM~\citep{yu2024siam}, a PoT-only method (we implement the two-stage version: SFT followed by DPO), to examine the impact of incorporating CoT into PoT reasoning. We implement these methods using the same source of data, critic models, and backbone model for fair comparisons.

As shown in Table~\ref{tab:main_results}, teaching LLMs to select the most suitable reasoning type for solving a given question remains a challenge. To construct the training data for DOTS, we designate PoT as the reasoning type if at least one PoT sample correctly answers a question while all CoT samples fail. The same principle applies in the reverse case. One possible explanation for this difficulty is the reliance on on-policy data to activate the internalized capabilities of LLMs. Currently, however, models primarily imitate the CoT/PoT capabilities of two teacher models (i.e., Qwen2-72B-Instruct and SIaM(Llama3-8B-Instruct)). In contrast, PoT-only methods like SiAM perform reasonably well, as they focus on distilling the Best-of-N distribution from a single PoT teacher model. When comparing CaP with SiAM, CaP can be seen as leveraging the previous CoT attempt to augment PoT in a multi-turn manner. Notably, CaP outperforms SiAM, regardless of the quality of the CoT used.

We observe that when using the same setting as CaP, but applying CoT to refine a prior CoT attempt, models fails to exhibit refinement behavior. For example, the performance remains unchanged (86.8\% vs 86.8\%) when self-generated CoT attempts are refined by Pair$_\text{SFT}$. Unlike CaP, preference optimization does not alleviate this issue, as shown by the results of Pair$_\text{DPO}$. Since the CoT ability of the CaP model after SFT improves by $3.5\%$, we regard the CoT attempts generated by the backbone model Qwen2-7B-Instruct as weak ones. Additionally, we observe a desirable trend of increasing PoT performance as CoT responses improve, with CaP$_\text{SFT}$ outperforming its initial CoT attempt when using tools. However, the improvement is marginal, and CaP$_\text{SFT}$ still struggles to effectively refine the CoT responses generated by Qwen2-72B-Instruct, which has stronger CoT capabilities. After preference optimization using CaP$_\text{SFT}$ as the reference policy, the performance gap before and after refinement widens significantly. For the first time, CaP (specifically, CaP$_\text{DPO}$) demonstrates the ability to effectively refine responses generated by Qwen2-72B-Instruct, despite being ten times smaller in size. Notably, with tools introduced, CaP appears capable of leveraging off-policy data, in contrast to previous CoT-CoT refinement studies that emphasized the necessity of self-generated data to avoid distribution mismatch~\citep{kumar2024training}.

\subsection{Scaling Test-Time Compute}
We further investigate the scaling of inference-time computation~\cite{brown2024large} when tools are utilized. In Figure~\ref{fig:combined}, SFT$_\text{PoT}$ and DPO$_\text{PoT}$ refer to PoT-only methods SiAM$_\text{SFT}$ and SiAM$_\text{DPO}$, respectively. In contrast to other variants and methods, CaP$_\text{DPO}$ further narrows the performance gap relative to Best-of-N sampling while using a smaller sample budget. Detailed results are provided in Table~\ref{tab:sampling}. Consistent with the trend observed under greedy decoding, using CoT responses from larger LLMs, such as CoT$_\text{72B}$, enhances final-turn performance -- especially when the sampling size is small. However, teacher LLMs may be unavailable at inference or could eventually be surpassed by self-improved LLMs. Rather than further enhancing the CoT capabilities of LLMs through post-training, we propose leveraging the internal capabilities of CaP models with a two-stage inference strategy, BoNBoN. First, we select the BoN CoT responses generated by the same LLM, and then sample PoT responses for an additional round of BoN selection. Experimental results show that reallocating compute to generate more CoT attempts within the same budget yields improved performance. While we adopt a balanced setting in this work, adaptive allocation strategies~\citep{manvi2024adaptive} -- guided by question difficulty and compute budgets -- may offer a promising direction for future studies.

\begin{table*}[h!]
\centering

\footnotesize
\begin{tabular}{lllllllllllll}
\toprule
model       & history & sampling & \multicolumn{2}{c}{Budget}  & \multicolumn{2}{c}{CM17K}  & APE & \multicolumn{2}{c}{CMATH} &  \multicolumn{2}{c}{average} \\
             & CoT             & strategy & CoT & PoT & valid & test & valid & valid & test &  \\
\midrule
SiAM$_\text{SFT}$   & -- &  greedy & 0 & 1 & 87.0 & 86.9 & 83.8 & 89.5 & 91.1 & 87.6 \\
\rowcolor{TableGreen}
                    & -- &  BoN  & 0 & 2 & 88.0  & 87.6  & 84.6 & 89.7 & 91.3 &  88.2 \\
\rowcolor{TableGreen}
                    &  &    &  & 8 & 89.5 & 89.9 & 87.7 & 91.5 & 94.2  & 90.5  \\
\rowcolor{TableGreen}
                    &  &    &  & 32 & 90.8 & 91.2 & 89.1 & 92.5 & 94.9 & 91.7 \\

SiAM$_\text{DPO}$    & -- &  greedy & 0 & 1 & 88.3 & 88.0 & 84.3 & 90.7 & 92.4 & 88.7 \\
\rowcolor{TableGreen}
                    & -- &  BoN & 0 & 2  & 88.4 & 87.9 & 85.3 & 91.0 & 93.0  & 89.1 \\
\rowcolor{TableGreen}
                    & -- &   & 0 & 8 & 89.6 & 89.7 & 86.8 & 92.5 & 93.4 & 90.4  \\
\rowcolor{TableGreen}
                     & -- &   & 0 & 32 & 90.8 & 91.3 & 88.2 & 93.3 & 95.4 & 91.8  \\

\midrule
CAP$_\text{SFT}$      & greedy(CoT$_\text{self}$)  & greedy & 1 & 1 & 86.9 & 87.1 & 84.3 & 89.8 & 91.1  & 87.8  \\ 
\rowcolor{TableGreen}
                      & greedy(CoT$_\text{self}$)    & BoN   & 1 & 2 & 88.7 & 89.0 & 85.3 & 88.3 & 91.5 & 88.6 \\
\rowcolor{TableGreen}
                      &    &    &  & 8  & 89.7 & 90.2 & 87.8 & 91.5 & 93.9 & 90.6 \\
\rowcolor{TableGreen}
                      &    &    &  & 32 & 90.8 & 91.3 & 89.4 & 92.8 & 95.8 &  92.0  \\

CAP$_\text{DPO}$      & greedy(CoT$_\text{self}$)  & greedy & 1 & 1 & 88.3 & 89.3 & 85.5 & 90.8 & 93.6 & 89.5\\
\rowcolor{TableGreen}
                      & greedy(CoT$_\text{self}$)    & BoN   & 1 & 2 & 88.9 & 89.6 & 86.2 & 91.7 & 93.8 & 90.0 \\
\rowcolor{TableGreen}
                      &    &    &  & 8 & 89.8 & 91.3 & 87.8 & 92.3 & 94.7 & 91.2 \\
\rowcolor{TableGreen}
                      &    &    &  & 32 & 90.8 & 91.9 & 89.3 & 92.8 & 95.5 & 92.1  \\
                      & BoN(CoT$_\text{self}$)   & BoNBoN  & 4 & 4  & 90.8 & 91.9 & 88.5 & 92.8 & 95.1 & 91.8\\
                      &    &  & 16 & 16  & 91.2 & 92.7 & 89.4 & 92.5 & 95.4 & 92.2\\
                      &    &  & 8 & 32  & 91.4 & 93.5 & 89.7 & 93.2 & 95.6 & 92.7\\
 \midrule
CAP$_\text{DPO}$       & greedy(CoT$_\text{72B}$) & greedy & 1 & 1  & 88.8 & 90.7 & 86.0 & 92.7 & 94.2 & 90.5\\
\rowcolor{TableGreen}
                       & greedy(CoT$_\text{72B}$) & BoN  & 1 & 2 & 89.3  & 90.6 & 86.7 & 92.2 & 94.2 & 90.6 \\
\rowcolor{TableGreen}
                       &  &   &  & 8 & 90.0  & 91.8 & 88.1 & 92.7 & 95.0 & 91.5 \\
\rowcolor{TableGreen}
                       &  &   &  & 32 & 91.3& 92.5 &89.3 & 92.0 & 95.9 & 92.2 \\

\bottomrule
\end{tabular}
\caption{Performance with increased test-time compute on out-of-distribution Chinese mathematical reasoning benchmarks.}

\label{tab:sampling}
\end{table*}

\subsection{Generalizability Across Other Backbone Models}

\begin{figure}[h]
\centering
    \begin{minipage}[b]{0.5\textwidth}
        \centering
        \footnotesize
\begin{tabular}{lcccc}
\toprule
model      &  stage   & first & refinement   \\
      &    & attempt &    \\
\midrule
tool         &  & \xmark  & \cmark  &    \\
\midrule
\multirow{2}{*}{Qwen2-7B-Instruct} & SFT        & 87.3 & 87.8 \tc{(+0.5)}  \\
                       &  DPO        & 87.2  & 89.5 \tc{(+2.3)} \\
\midrule
\multirow{2}{*}{Qwen2.5-7B-Instruct} & SFT         & 88.5 & 89.0 \tc{(+0.5)}\\
                       &  DPO      & 88.5 & 90.3 \tc{(+1.8)} \\
\midrule
\multirow{2}{*}{Llama3-8B-Instruct} & SFT       & 82.9 &  86.6 \tc{(+3.7)}   \\
                       & DPO      & 82.5 & 87.9  \tc{(+5.4)} \\
\bottomrule
\end{tabular}
\caption{Average accuracy comparison of CaP SFT/DPO models trained with different backbone models, using greedy decoding during inference.}
\label{tab:exp:backbone}
    \end{minipage}
    \hfill
    \begin{minipage}[b]{0.47\textwidth}
        \centering
        \includegraphics[width=\textwidth]{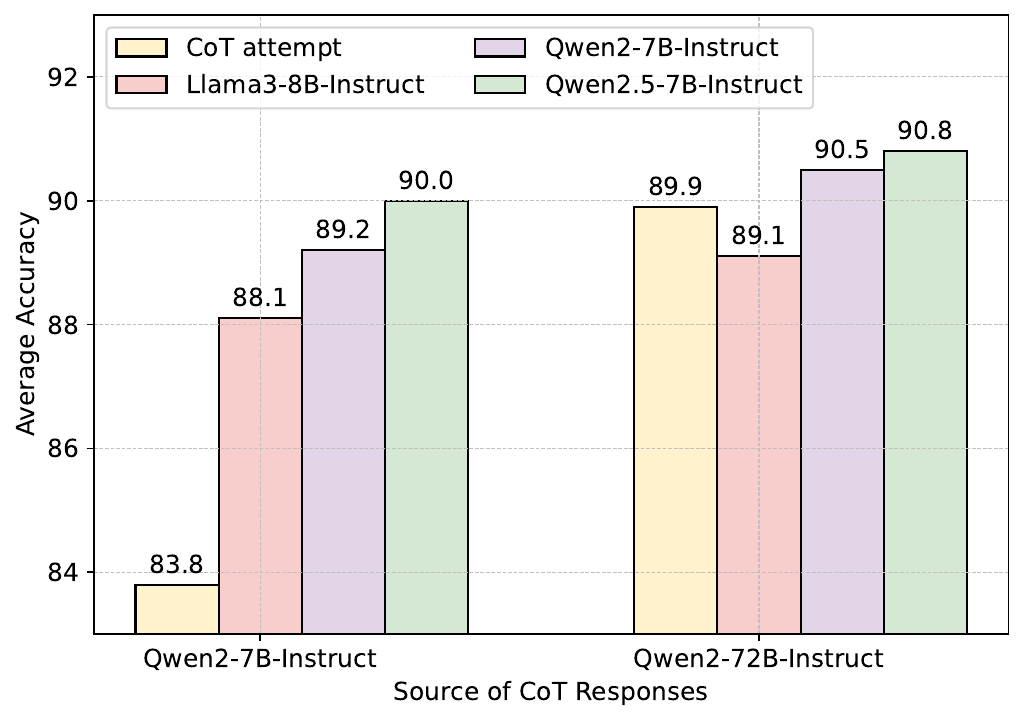} 
        \caption{Refining off-policy responses using CaP$_\text{DPO}$ with three backbone models.}
        \label{fig:exp:backbone:offpolicy}
    \end{minipage}
\end{figure}

Additionally, we investigate other models, including Llama3-8B~\citep{llama3} and Qwen2.5-7B-Instruct~\citep{yang2024qwen25}, which exhibit worse or better performance on mathematical benchmarks, to evaluate the generalization ability of our method. As shown in Figure~\ref{tab:exp:backbone}, all three models trained with CaP can achieve better performance after self-refinement. 

It is worth noting that the CoT reasoning ability of Llama3-8B-Instruct in solving Chinese mathematical questions is relatively weaker compared to other models that are extensively pre-trained on Chinese corpora. For CaP(Llama3-8B-Instruct), CoT responses generated by Qwen2-7B-Instruct exhibit fewer errors than its self-generated CoT attempts (83.8\% vs. 82.5\%), as shown in Figure~\ref{fig:exp:backbone:offpolicy}. Nevertheless, CaP(Llama3-8B-Instruct) is still capable of refining responses from Qwen2-7B-Instruct, improving their accuracy from 83.8\% to 88.1\%. However, it struggles to refine responses when there is a significant disparity in Chinese CoT reasoning capabilities between the two LLMs (e.g., Llama3-8B-Instruct vs. Qwen2-72B-Instruct). This observation suggests that an LLM's ability to refine responses may be closely tied to its own problem-solving proficiency.

\subsection{Discussions and Remaining Challenges}

This study primarily focuses on refining a single response, but CaP can be easily adapted for multi-attempt refinement by extending the context to include multiple responses along with their corresponding critics.
In preliminary experiments, we explore a two-attempt setting structured in a linear sequence while keeping other factors unchanged. However, despite incurring additional costs for CoT sampling and ranking, the refinement performance shows only a marginal improvement of 0.3\% compared to the one-attempt CaP with the default positive critic. Since BoNBoN already involves CoT response ranking, we leave further exploration of context extension for future work.

Moreover, the question of how to refine PoT responses using CoT, enabling reasoning and refinement with alternating reasoning types, remains unresolved. As shown in Table~\ref{fig:exp:alternating} with Qwen2.5-7B-Instruct, simply introducing an additional PoT-CoT refinement task during the supervised fine-tuning stage of CaP models negatively impacts both CoT reasoning and PoT refinement capabilities.

\begin{table}[h!]
\centering

\footnotesize
\begin{tabular}{lll}
\toprule
Tasks &  first attempt & refinement  \\
              &(Q $\rightarrow$ CoT)    & (Q,CoT$\rightarrow$ PoT)  \\
\midrule
Q $\rightarrow$ CoT; Q,CoT$\rightarrow$ PoT & 88.5 & 89.0 \\
Q $\rightarrow$ CoT; Q,CoT$\rightarrow$ PoT; Q, PoT$\rightarrow$ CoT &  88.1 \tr{(-0.4)} & 88.4 \tr{(-0.6)}\\
\bottomrule
\end{tabular}
\caption{Performance drop when a new refinement task is introduced.}
\label{fig:exp:alternating}
\end{table}

Introducing tools for refinement can enhance the robustness of LLMs. For example, we observe a 1.0\% drop in average accuracy for Qwen2-72B-Instruct, a state-of-the-art CoT-style LLM, simply by adding the word ``please'' before each question. In contrast, PoT-based methods remain nearly unaffected, showing the vulnerability of CoT reasoning to subtle linguistic variations when stepwise rewards or critics are unavailable. Adopting a more concise and precise programming language may help alleviate this issue, improving consistency in performance. Ultimately, such refinements have the potential to enable deeper reasoning trajectories while ensuring controlled and reliable quality.

\section{Conclusions and Future Work}

We introduce CaP, a new method that leverages external tools to refine chain-of-thought responses generated by the same or different LLMs. CaP utilizes a two-phase training strategy: supervised fine-tuning followed by preference optimization using variants of DPO. Our findings emphasize the critical role of preference optimization in achieving effective refinement. Furthermore, we explore multiple sampling strategies to integrate both CoT and PoT during inference. Experimental results showcase CaP's ability to facilitate cross-reasoning refinement and efficient inference.

Future work includes training and evaluating CaP in multilingual settings, as well as enhancing its capabilities through adaptive allocation strategies, online alignment algorithms, active learning with human-in-the-loop, and the development of more expressive error-aware critic models that focus on either processes or outcomes. Additionally, CaP models have the potential to serve as a step-level translator between natural language and programming languages, leveraging its training on solution-level ``parallel'' CoT-PoT data.

\bibliography{neurips_2024}
\bibliographystyle{neurips_2024}

\clearpage
\newpage
\appendix

\section{Appendices}
\label{sec:appendix}

\begin{table*}[ht!]

\centering
\footnotesize

\begin{tabular}{lp{11cm}}
\toprule

\textbf{Instruction} & Your goal is to evaluate whether the model's candidate answer to the given math problem is correct.  \\
    &  -- First, carefully read the math problem.\\
    &  -- Then, check the candidate answer provided by the model, which may be a step-by-step reasoning process or a piece of code. \\
    &  -- Finally, conclude whether the candidate answer is correct or not. \\
    &  -- The response format should be Yes or No.\\
\midrule

\textbf{Input}    & \#\#\# Question \\
                        & A cylinder is cut in half along its height through the diameter, increasing the surface area by 60 square centimeters. Find the lateral surface area of the cylinder. \\
                        & \#\#\# Candidate Answer \\

& \begin{lstlisting}[basicstyle=\ttfamily\scriptsize, breaklines=true, aboveskip=0pt, belowskip=0pt]
Let's solve this step by step.

First, we know that the increase in the cylinder's surface area is due to the addition of a rectangular face on each of the two half-cylinders after the cylinder is cut in half. The length of this rectangle is the height of the cylinder, and its width is the diameter of the cylinder's base.

Since the surface area increases by 60 square centimeters, the area of each rectangle is half of this, which is 60 divided by 2, resulting in 30 square centimeters.

Let the height of the cylinder be "h" and the diameter of its base be "d". The area of the rectangle is then given by d multiplied by h, which means d * h = 30.

The lateral surface area of the cylinder is calculated as the height multiplied by the circumference of the base, which is h * pi * d.

Since d * h = 30, we can express d as 30 divided by h.

Substituting d = 30 / h into the formula for lateral surface area, we get:
S = h * pi * (30 / h).

Simplifying this, we find:
S = 30 * pi.

Answer: The lateral surface area of the cylinder is 30 * pi square centimeters.
\end{lstlisting} \\
& \#\#\# Assessment \\

\midrule
\textbf{Output} & Yes \\

\bottomrule
\end{tabular}
\caption{Example instances of reference-free critic data (CoT).}
\label{tab:reference_free_critic_cot}
\end{table*}

\begin{table*}[ht!]

\centering
\footnotesize

\begin{tabular}{lp{11cm}}
\toprule

\textbf{Instruction} & Your goal is to evaluate whether the model's candidate answer to the given math problem is correct.  \\
    &  -- First, carefully read the math problem.\\
    &  -- Then, check the candidate answer provided by the model, which may be a step-by-step reasoning process or a piece of code. \\
    &  -- Finally, conclude whether the candidate answer is correct or not. \\
    &  -- The response format should be Yes or No.\\
\midrule

\textbf{Input}    & \#\#\# Question \\
                        & The school evenly distributed 135 exercise books to 3 classes. How many books does each class get? \\
                        & \#\#\# Candidate Answer \\
& \begin{lstlisting}[basicstyle=\ttfamily\scriptsize, breaklines=true, aboveskip=0pt, belowskip=0pt]
# Based on the problem-solving process, calculate the number of exercise books each class gets

# Total number of exercise books
total_books = 135
# Number of classes
num_classes = 3

# Calculate the number of exercise books per class
books_per_class = total_books / num_classes
books_per_class
\end{lstlisting} \\
                        & \#\#\# Assessment \\
\midrule
\textbf{Output} & Yes \\

\bottomrule
\end{tabular}
\caption{Example instances of reference-free critic data (PoT).}
\label{tab:reference_free_critic_pot}
\end{table*}

\begin{table*}[ht!]

\centering
\footnotesize

\begin{tabular}{lp{11cm}}
\toprule

\textbf{Instruction} & Your goal is to evaluate whether the candidate answer provided by the model for a math problem is consistent with the reference answer. The steps to complete this task are as follows.  \\
    &  -- First, carefully read the given math problem. \\
    &  -- Next, review the reference answer for the math problem. \\
    &  -- Then, examine the candidate answer provided by the model, which may include a program as well as the result of running the program. \\
    &  -- Finally, determine whether the candidate answer is consistent with the reference answer or can be made consistent through simple calculations/conversions.\\
    & -- The response format should be: Yes or No. \\
\midrule

\textbf{Input}    & \begin{lstlisting}[basicstyle=\ttfamily\scriptsize, breaklines=true, aboveskip=0pt, belowskip=0pt]
### Question
The range of the function $y=\sqrt{x^2-2x-3}$ is.
### Reference Answer
$[0, +\infty)$.
### Candidate Answer
<code>
# Define a function to calculate the nth term
def calculate_nth_term(n):
    # Using the formula derived from analysis
    an = 2 ** n - 2 ** (n - 1)
    return an

# Calculate the value of the 3rd term
a3 = calculate_nth_term(3)

# Return the result as a dictionary
result = {'a3': a3}
result

</code>
<solution>{'a3': 4}</solution>
### Assessment
\end{lstlisting} \\
\midrule
\textbf{Output} & No \\

\bottomrule
\end{tabular}
\caption{Example instances of reference-based critic data (PoT).}
\label{tab:reference_based_critic_pot}
\end{table*}

\end{document}